%% file: main.tex
\documentclass[10pt,twocolumn,letterpaper]{article}
\usepackage{cvpr}
\usepackage[pagebackref,breaklinks,colorlinks,citecolor=cvprblue]{hyperref}
\usepackage[accsupp]{axessibility}

\input{preamble}

\input{custom}

\input{mathabbr}

\definecolor{cvprblue}{rgb}{0.21,0.49,0.74}
\def\paperID{****}
\def\confName{CVPR}
\def\confYear{2024}

\title{Synergistic Global-space Camera and Human Reconstruction from Videos}

\author{
Yizhou Zhao$^{1}$\thanks{},
Tuanfeng Yang Wang$^{2}$,
Bhiksha Raj$^{1}$,
Min Xu$^{1}$,
Jimei Yang$^{2}$,
Chun-Hao Paul Huang$^{2}$ \\
$^{1}$Carnegie Mellon University \quad\quad
$^{2}$Adobe Research \\
{\tt\small \{yizhouz,bhiksha,mxu1\}@cs.cmu.edu, \{yangtwan,jimyang,chunhaoh\}@adobe.com}
}

\begin{document}
\twocolumn[{%
\renewcommand\twocolumn[1][]{#1}%
\maketitle
\input{sections/0_teaser}
}]
\renewcommand{\thefootnote}{\fnsymbol{footnote}}
\footnotetext[1]{Part of this work was done when interned at Adobe Research.}
\input{sections/0_abstract}    
\input{sections/1_intro}
\input{sections/2_related_work}
\input{sections/3_method}
\input{sections/4_experiments}
\input{sections/5_limitation}
\input{sections/6_conclusion}
\input{sections/7_acknowledgment}

{
    \small
    \bibliographystyle{ieeenat_fullname}

\input{main.bbl}
}
\input{sections/X_suppl}

\end{document}

%% file: preamble.tex
\usepackage[dvipsnames]{xcolor}


%% file: custom.tex
\usepackage{adjustbox}
\usepackage{bm}
\usepackage{multirow}
\usepackage{multicol}
\usepackage{color, colortbl}
\usepackage{pifont}

\usepackage[capitalize]{cleveref}
\Crefname{section}{Section}{Sections}
\crefname{section}{Sec.}{Secs.}
\Crefname{align}{Equation}{Equations}
\crefname{align}{Eq.}{Eqs.}
\Crefname{equation}{Equation}{Equations}
\crefname{equation}{Eq.}{Eqs.}
\Crefname{figure}{Figure}{Figures}
\crefname{figure}{Fig.}{Figs.}
\Crefname{table}{Table}{Tables}
\crefname{table}{Tab.}{Tabs.}

\newcommand\minisection[1]{\vspace{1mm}\noindent \textbf{#1}}
\newcommand{\norm}[1]{\left\lVert#1\right\rVert}

\newcommand{\cmark}{\ding{51}}
\newcommand{\xmark}{\ding{55}}
\newcommand{\nameCOLOR}[1]{\textcolor{black}{#1}} 
\newcommand{\nameMethod}{\nameCOLOR{\mbox{SynCHMR}}\xspace}
\newcommand{\nameMethodFull}{\nameCOLOR{Synergistic Camera and Human Reconstruction}\xspace}
\newcommand{\synchmr}{\nameMethod}
\newcommand{\hmr}{\nameCOLOR{\mbox{HMR}}\xspace}
\newcommand{\camspacehmr}{\nameCOLOR{camera-frame HMR}\xspace}
\newcommand{\worldspacehmr}{\nameCOLOR{world-frame HMR}\xspace}

\newcommand{\slam}{\nameCOLOR{\mbox{SLAM}}\xspace}
\newcommand{\phalpp}{\nameCOLOR{\mbox{PHALP$^+$}}\xspace}
\newcommand{\slahmr}{\nameCOLOR{\mbox{SLAHMR}}\xspace}
\newcommand{\fourdhumans}{\nameCOLOR{\mbox{4DHumans}}\xspace}
\newcommand{\threedpw}{\nameCOLOR{\mbox{3DPW}}\xspace}
\newcommand{\egobody}{\nameCOLOR{\mbox{EgoBody}}\xspace}
\newcommand{\zoedepthplus}{\nameCOLOR{\mbox{ZoeDepth$^+$}}\xspace}
\newcommand{\supp}{\mbox{Supp.~Mat.}\xspace}

\definecolor{Gray}{gray}{0.9}
\definecolor{Celadon}{rgb}{0.67, 0.88, 0.69}
\definecolor{Cream}{rgb}{1.0, 0.99, 0.82}

\DeclareMathOperator{\diag}{diag}
\DeclareMathOperator{\argmin}{argmin}

%% file: mathabbr.tex
\newcommand{\Cc}{\mathcal{C}}
\newcommand{\Dc}{\mathcal{D}}
\newcommand{\Ec}{\mathcal{E}}

\newcommand{\Lc}{\mathcal{L}}

\newcommand{\Pc}{\mathcal{P}}

\newcommand{\Rb}{\mathbb{R}}

\newcommand{\cv}{\mathbf{c}}
\newcommand{\dv}{\mathbf{d}}

\newcommand{\pv}{\mathbf{p}}

\newcommand{\rv}{\mathbf{r}}

\newcommand{\tv}{\mathbf{t}}

\newcommand{\wv}{\mathbf{w}}
\newcommand{\xv}{\mathbf{x}}

\newcommand{\zv}{\mathbf{z}}

\newcommand{\Dv}{\mathbf{D}}

\newcommand{\Gv}{\mathbf{G}}

\newcommand{\Iv}{\mathbf{I}}
\newcommand{\Jv}{\mathbf{J}}

\newcommand{\Mv}{\mathbf{M}}

\newcommand{\Pv}{\mathbf{P}}

\newcommand{\Rv}{\mathbf{R}}
\newcommand{\Sv}{\mathbf{S}}

\newcommand{\Vv}{\mathbf{V}}

\newcommand{\Xv}{\mathbf{X}}
\newcommand{\Yv}{\mathbf{Y}}

\newcommand{\betav      }{\boldsymbol \beta      }

\newcommand{\thetav     }{\boldsymbol \theta     }

\newcommand{\Gammav     }{\boldsymbol \Gamma     }

\newcommand{\Phiv       }{\boldsymbol \Phi       }

%% file: sections/0_teaser.tex
\begin{center}
    \centering
    \vspace{-2.5em}
    \includegraphics[width=0.96\linewidth]{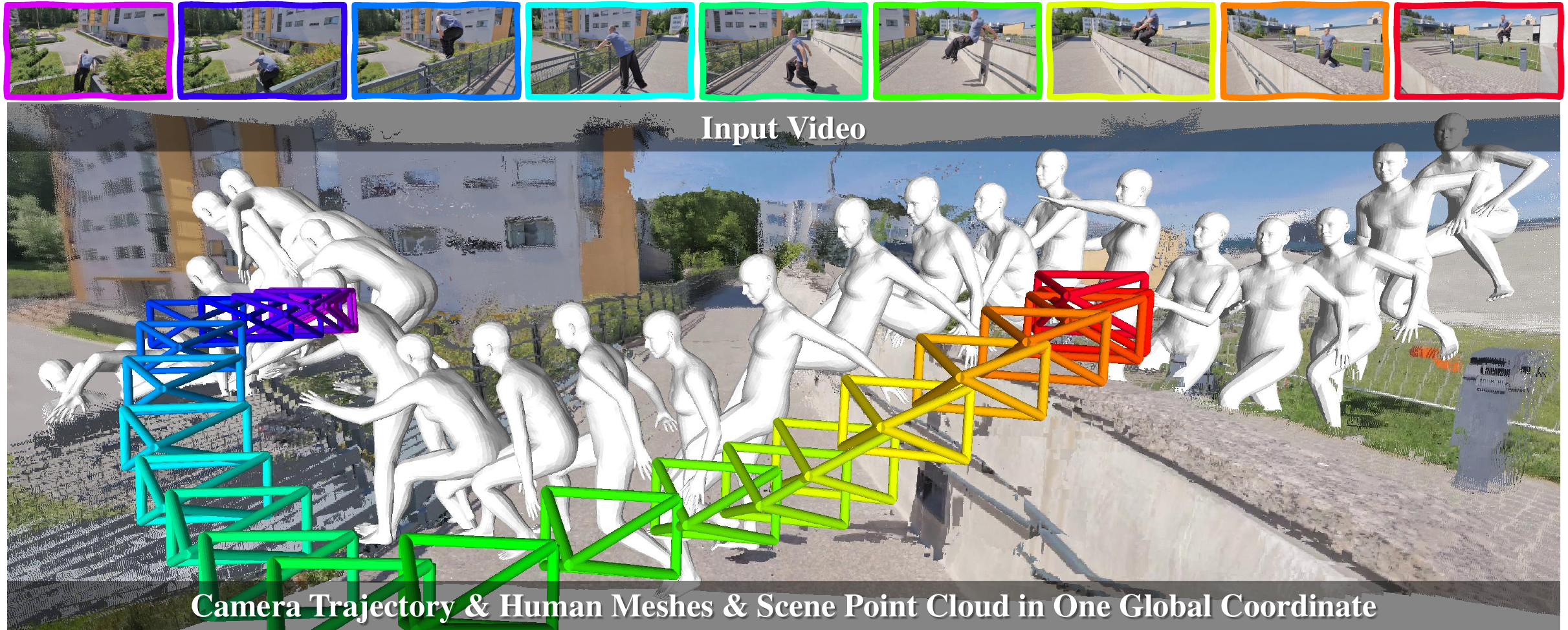}
    \vspace{-0.8em}
    \captionsetup{type=figure}
    \caption{Our \nameMethod recovers metric-scale camera trajectories (color pyramids), human meshes, and scene point clouds in a global coordinate from casual videos by joining forces of Human Mesh Recovery (HMR) and Simultaneous Localization and Mapping (SLAM).}
    \label{fig:teaser}
\end{center}%

%% file: sections/0_abstract.tex
\begin{abstract}
\vspace{-0.6em}
Remarkable strides have been made in reconstructing static scenes or human bodies from monocular videos.
Yet, the two problems have largely been approached independently, without much synergy.
Most visual \slam methods can only reconstruct camera trajectories and scene structures up to scale, while most \hmr methods reconstruct human meshes in metric scale but fall short in reasoning with cameras and scenes.
This work introduces \nameMethodFull (\nameMethod) to marry the best of both worlds.
Specifically, we design Human-aware Metric SLAM to reconstruct metric-scale camera poses and scene point clouds using \camspacehmr as a strong prior, addressing depth, scale, and dynamic ambiguities.
Conditioning on the dense scene recovered, we further learn a Scene-aware SMPL Denoiser to enhance \worldspacehmr by incorporating spatio-temporal coherency and dynamic scene constraints.
Together, they lead to consistent reconstructions of camera trajectories, human meshes, and dense scene point clouds in a common world frame. Project page: {\small \url{https://paulchhuang.github.io/synchmr}}
\end{abstract}

%% file: sections/1_intro.tex
\section{Introduction}
\label{sec:intro}
\vspace{-0.6em}

\begin{figure}[ht]
    \begin{center}
        \includegraphics[width=\linewidth]{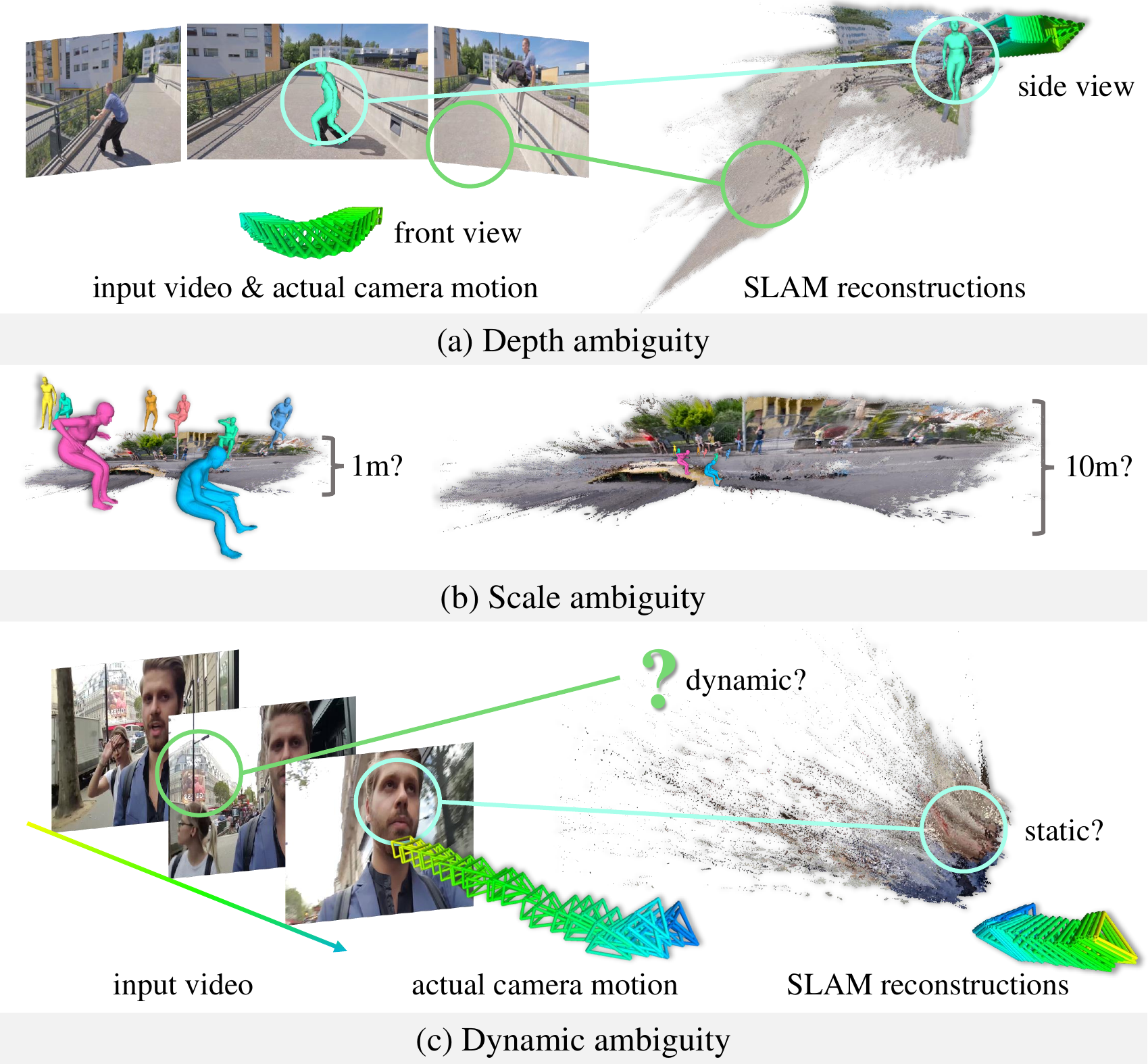}
    \end{center}
    \vspace{-1.5em}
    \caption{\textbf{Illustration of three types of ambiguities in visual SLAM.} We show SLAM reconstruction results from DROID-SLAM~\cite{teed2021droid}. (a) Depth ambiguity occurs when there are only minor camera translations between different views. This can lead to geometric failures in reconstruction such as the folded back corridor in the side view. (b) Scale ambiguity is inherent in monocular SLAM systems and requires additional reference for disambiguation. (c) Dynamic ambiguity gets pronounced when moving foregrounds dominate frames. Over-reliance on foreground key points will result in incorrect camera trajectories.}
    \label{fig:ambiguity}
    \vspace{-1.5em}
\end{figure}

Physically plausible 3D human motion reconstruction from monocular videos is a long-standing problem in computer vision and graphics and has many applications in character animation, VFX, video games, sports, and healthcare. 
It requires estimating 3D humans across video frames in a common coordinate even with a moving camera.
While human mesh recovery (\hmr) has made significant progress recently \cite{tian2022hmrsurvey},
most existing methods typically estimate 3D humans in the camera coordinate by one frame at a time and fail to disambiguate camera motion.
It calls for methods to jointly reconstruct 3D human and camera motion in a consistent global coordinate system from monocular videos.
In other words, taking a video captured by a moving camera as input, the method should recover both temporally and spatially coherent movements of human bodies and cameras.

Intuitively, if the accurate camera motion is given, one can transform the bodies from individual camera frames to a common world frame by multiplying the inverse of camera extrinsic matrices. 
In practice, with humans moving in the scene, estimating the camera motion of a video is still an open challenge in monocular \slam~\cite{slam2022survery}.
It not only falls short in capturing accurate depths on views with small camera translations but more crucially, only estimates scene structures and camera trajectories \emph{up to scale}.
The human motion also breaks the static key point assumption in the bundle adjustment.
As a result, one needs additional reference to disambiguate the depth, the scale, and the dynamic as illustrated in~\cref{fig:ambiguity}.

To leverage \slam results in \hmr pipelines, current \worldspacehmr methods often refine camera poses by integrating either partial camera parameters, such as a global scale of the translation~\cite{ye2023slahmr}, or full extrinsic matrices~\cite{saini2023smartmocap,henning2022bodyslam,kocabas2024pace} in an optimization-based framework.
However, their optimization-based nature leads to complex multi-stage schemes, making the overall pipeline unnecessarily slow and easy to break.

In this work, we explore a fundamentally different way to marry the best of \hmr and \slam.
A 2D object can first be lifted from the image plane to the camera frame and then transformed into a common 3D space. 
This two-step process coincides with the combination of \camspacehmr, which brings imaged 2D humans to 3D camera frames, and \slam, which estimates the camera-to-world transformation.
Noticing these correspondences, we leverage \camspacehmr as a strong prior to bridge from the image plane to the camera frame for disambiguating \slam, and utilize \slam reconstructions to constrain the transformation of human meshes from individual camera frames to a common global space.
The overall pipeline thus results in a better synergy of the two, which we dub \nameMethodFull (\nameMethod).

We design \nameMethod based on several insights.
First, despite \camspacehmr methods cannot reconstruct humans in a coherent global frame, the estimated body dimension and location still provide cues to disambiguate \slam.
Unlike \slahmr~\cite{ye2023slahmr} which applies \slam out of the box and corrects the scale afterward, we endow the \slam process with human meshes from \camspacehmr to address ambiguities.
To this end, we capitalize on estimated \emph{absolute} depths to provide pseudo-RGB-D inputs for \slam~\cite{teed2021droid} and confine the bundle adjustment to static backgrounds.
Since current depth estimation methods~\cite{bhat2023zoedepth,ranftl2020towards} predicts either relative depth maps or depths with data biases, we propose to calibrate their outputs by aligning with estimated human bodies in the camera frame~\cite{goel2023humans}.
With these priors, \slam knows the depth, scale, and dynamic information from \hmr and consequently estimates less ambiguous scene structures and camera poses.

Next, we place human meshes in the common coordinate recovered by \slam.
The gap between human tracks transformed from camera frames and their real plausible world-frame motions stems from two sources of error: noise induced by \camspacehmr and by \slam. 
Motion prior models \cite{rempe2021humor,he2022nemf,Zhang_2021_ICCV} can be used for denoising purposes as they contribute to the temporal coherence of world-frame human tracks.
However, their exclusive focus on human modeling either leaves them agnostic to the underlying scenes \cite{he2022nemf} or assumes the scene is a simple ground plane \cite{rempe2021humor,Zhang_2021_ICCV}.
Our intuition is that when placing a human, static elements of the scene, such as the ground, and dynamic components like moving objects are both possible to be in contact with the human, thereby providing clues for placing the body coherently and compatibly with the scene.
We hence introduce a Scene-aware SMPL Denoiser that learns to denoise the transformed human tracks by considering both temporal consistencies of moving humans and implicit constraints from dynamic scenes.
This global awareness makes it more flexible for in-the-wild videos.

Our contributions can be summed up as follows:
\begin{itemize}
    \item We present a novel pipeline, \nameMethod, that takes a monocular video as input and reconstructs human motions, camera trajectory and \textit{dense} scene point clouds all in one global coordinate, as shown in~\cref{fig:teaser}, whereas current \worldspacehmr methods \cite{ye2023slahmr,kocabas2024pace,yuan2021glamr} can recover only an estimated or pre-defined ground plane.
    \item We propose a novel Human-aware Metric SLAM process to robustly calibrate estimated depth with estimated human meshes, resulting in metric-scale camera pose estimation and metric-scale scene reconstruction.
    \item We present Scene-aware SMPL Denoising that enforces spatiotemporal coherencies and applies dynamic scene constraints on world-frame human meshes. Notably, this is achieved without requiring extra annotations or heuristic designs to decide which part of a human should be interacting with the scene~\cite{shen2023learning} and which region in the scene is most likely to be in contact with humans~\cite{luvizon2023samhmc,ye2023slahmr}.
\end{itemize}

%% file: sections/2_related_work.tex
\begin{figure*}[ht]
    \begin{center}
        \includegraphics[width=\linewidth]{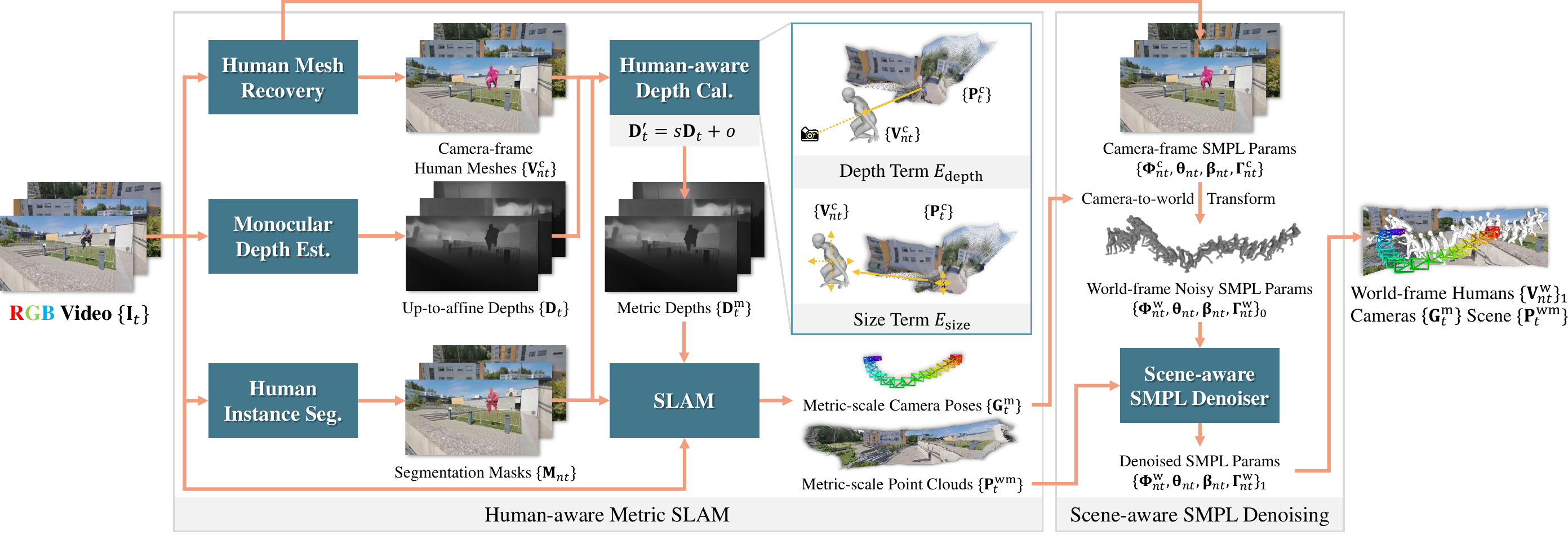}
    \end{center}
    \vspace{-1.0em}
    \caption{\textbf{The architecture of \nameMethod.} Our pipeline comprises two phases. The first phase, Human-aware Metric SLAM (\cref{subsec:human-aware-slam}), infers metric-scale camera poses and metric-scale point clouds by exploiting the camera-frame human prior. The second phase, Scene-aware SMPL Denoising (\cref{subsec:smpl_denoising}), involves the conditional denoising of world-frame noisy SMPL parameters. These parameters, initialized by transforming from the camera frame, get refined through conditioning on the dynamic point clouds obtained in the first phase. The whole pipeline thus reconstructs humans, scene point clouds, and cameras harmoniously in a common world frame.}
    \label{fig:pipeline}
\vspace{-1.0em}
\end{figure*}

\section{Related Work}
\label{sec:related}
There is considerable prior arts of \hmr. We briefly discuss how they adopt different camera models and refer the readers to \cite{slam2022survery} for a more comprehensive review.  

\minisection{\hmr from a single image.}
State-of-the-art (SOTA) methods use parametric body models \cite{totalcapture,SMPL:2015,SMPL-X:2019,xu2020ghum} 
and estimate the parameters either by fitting to detected image features \cite{bogo_smplify,SMPL-X:2019,xiang2019monocular} or by regressing directly from pixels with deep neural networks \cite{PARE_2021,guler_2019_CVPR,jiang2020mpshape,joo2021eft,kanazawa_hmr,SPIN:ICCV:2019,rong2020frankmocap,song2020human,xu2019denserac,zanfir2020weakly,pymaf2021,goel2023humans,Lin_2023_CVPR,li2021hybrik}.
These approaches
assume weak perspective/orthographic projection or pre-define the focal length as a large constant for all images. 
Kissos \etal~\cite{kissosECCVW2020} show that replacing focal length with a constant closer to ground truth alleviate the body tilting problem.
SPEC \cite{Kocabas_SPEC_2021} and Zolly \cite{wang2023zolly} estimate focal length to account for perspective distortion.
CLIFF \cite{li2022cliff} takes into account the location of humans in images to regress better poses in the camera coordinates.

Many of these \camspacehmr methods assume zero camera rotation, which entangles body rotation and camera rotation.
When applied on video data, they fail to reconstruct humans in a coherent global space since they operate in a per-frame manner and hence cannot reason about how the camera moves across frames.

\noindent\textbf{\hmr from videos} aims to regress a series of body parameters from a temporal sequence. 
It opens up new problems such as whether the reconstructed bodies are in a common global coordinate or not.
Some temporal methods consider a static camera \cite{rempe2021humor,Zhang_2021_ICCV,luvizon2023samhmc}, which makes the camera space a natural choice of the common coordinate.
The challenge of coherent global space emerges when the camera moves.
Early methods \cite{humanMotionKanazawa19, kocabas2019vibe, choi2020beyond} show promising results on videos of dynamic cameras. 
Despite the reconstructed human meshes look great when overlaid on images, they do not share a common coordinate in 3D. 

Recent \hmr methods capitalize on human motion prior to constrain the global trajectories in the world space, which in turn implicitly disentangles human movement from camera movement.
GLAMR \cite{yuan2021glamr} consider a data-driven prior models learned on large-scale MoCap database \eg~AMASS \cite{AMASS_2019}, while D\&D \cite{li2022dnd} and Yu \cite{yu2021movingcam} consider physic-inspired prior. 
These \worldspacehmr methods often struggle on noise in local poses caused by partial occlusions, which is very common in in-the-wild videos with close-up shots and crowd scenes.
Kaufmann \etal \cite{kaufmann2023emdb} and BodySLAM++ \cite{henning2023bodyslam++} circumvent this problem by employing IMU sensors to provide more robust body estimates but require extra sensory devices.
To fully disentangle human and camera motion, 
another line of work~\cite{liu20214d,saini2023smartmocap,henning2022bodyslam,kocabas2024pace} leverages state-of-the-art \slam techniques, \eg~\cite{zhang2022casualsam,teed2021droid,schoenberger2016sfm}, to explicitly estimate camera motion from the input video and infer the body parameters in the world coordinate of \slam.
Closest to us is \slahmr \cite{ye2023slahmr} which solves for a global scale to connect the pre-computed \slam results and body trajectories.
To carefully guide the optimization process, these methods tend to have complex, multi-stage optimization schemes, making the overall pipeline easy to break and unnecessarily slow.

Note that in stark contrast to the methods above, which either assume or estimate a simple ground plane as scene representation, 
\nameMethod reconstructs dense scenes from in-the-wild videos without pre-scanning with extra devices \textit{a priori} like in \cite{zhang2022egobody, Hassan2019prox, huang2022rich, guzov2021hps, Dai_2022_CVPR, Dai_2023_CVPR, yan2023cimi4d}.
We provide detailed comparisons with these \worldspacehmr in \supp

%% file: sections/3_method.tex
\section{Method}
\label{sec:method}
Taking as input an RGB video $\{\Iv_t \in \Rb^{H \times W \times 3}\}_{t=1}^T$ with $T$ frames and $N$ people in the scene, we aim to recover human meshes $\{\Vv_{nt}^\text{w} \in \Rb^{3 \times 6890}\}_{n=1,t=1}^{N,T}$, dynamic scene point clouds $\{\Pv_t^\text{wm} \in \Rb^{H \times W \times 3} \}_{t=1}^T$, and corresponding camera poses $\{\Gv_t^\text{m} \in \text{SE}(3)\}_{t=1}^T$ in a common world coordinate system. The superscripts $^\text{w}$, $^\text{c}$, and $^\text{m}$ denote the world frame, the camera frame, and the metric scale, respectively. 
To this aim, we propose a two-phase alternative conditioning pipeline as depicted in~\cref{fig:pipeline}. In the first phase, we calibrate camera motion by injecting a camera-frame human prior to \slam. This resolves depth, scale, and dynamic ambiguities, yielding metric-scale camera poses and dynamic point clouds. Subsequently, in the second phase, we transform the camera-frame human tracks into the world frame and utilize the dynamic point clouds obtained in the first phase for conditional denoising.

\subsection{Preliminaries}
\subsubsection{SLAM}
Given a monocular RGB video $\{\Iv_t\}_{t=1}^T$, DROID-SLAM~\cite{teed2021droid} solves a dense bundle adjustment for a set of camera poses $\{\Gv_t \in \text{SE}(3)\}_{t=1}^T$ and inverse depths $\{\dv_t\in \Rb^{H \times W}_+\}_{t=1}^T$. To update these estimations, it first computes a dense correspondence field $\pv_{ij} \in \Rb^{H\times W \times 2}$ based on reprojection for each pair of frames $(i, j)$:
\begin{align}
    \pv_{ij} = \Pi(\Gv_{ij} \circ \Pi^{-1}(\pv_i, \frac{1}{\dv_i})),
\end{align}
where $\pv_i \in \Rb^{H \times W \times 2}$ is a grid of pixel coordinates in frame $i$, $\Gv_{ij} = \Gv_j \circ \Gv_i^{-1}$ is the relative pose, and $\Pi$ and $\Pi^{-1}$ are the camera projection and inverse projection functions. Then with a learned neural network, the system predicts a revision flow field $\rv_{ij} \in \Rb^{H \times W \times 2}$ and associated confidence map $\wv_{ij} \in \Rb_+^{H \times W \times 2}$ to construct the cost function
\begin{align}
    E_\Sigma = \sum_{(i,j)} \norm{\pv_{ij}^* - \Pi(\Gv'_{ij} \circ \Pi^{-1}(\pv_i, \frac{1}{\dv'_i})) }_{\Sigma_{ij}}^2,
\label{eq:slam}
\end{align}
where $\pv_{ij}^* = \rv_{ij} + \pv_{ij}$ is the corrected correspondence, $\norm{\cdot}_{\Sigma}$ is the Mahalanobis distance which weighs the error terms with $\Sigma_{ij} = \diag \wv_{ij}$, and $\Gv'$ and $\dv'$ are updated poses and inverse depths. Upon this objective, DROID-SLAM considers an additional term that penalizes the squared distance between the measured and predicted depth if the input is with an extra sensor depth channel $\{\Dv_t\}_{t=1}^T$.

\vspace{-0.5em}
\subsubsection{HMR}
We employ 4DHumans~\cite{goel2023humans} for reconstructing camera-frame human meshes from an in-the-wild video. Specifically, it performs per-frame human mesh recovery with an end-to-end transformer architecture and associates them to form human tracks. Each tracked human $n$ in frame $t$ is represented by SMPL~\cite{SMPL:2015} parameters as $\{\bm{\Phi}_{nt}, \bm{\theta}_{nt}, \bm{\beta}_{nt}, \bm{\Gamma}_{nt}\}$, including global orientation $\bm{\Phi}_{nt} \in \Rb^{3 \times 3}$, body pose $\bm{\theta}_{nt} \in \Rb^{22 \times 3 \times 3}$, shape $\bm{\beta}_{nt} \in \Rb^{10}$, and root translation $\bm{\Gamma}_{nt} \in \Rb^3$. Then the parametric SMPL model can use these parameters to recover a human mesh with vertices $\Vv_{nt} \in \Rb^{3 \times 6890}$ in metric scale: $\Vv_{nt} = \text{SMPL}(\bm{\Phi}_{nt}, \bm{\theta}_{nt}, \bm{\beta}_{nt}) + \bm{\Gamma}_{nt}$.

\subsection{Human-aware Metric SLAM}\label{subsec:human-aware-slam}
\subsubsection{Preprocessing}
To start off, we estimate per-frame depth maps $\{\Dv_t\}$ with an off-the-shelf depth estimator, ZoeDepth~\cite{bhat2023zoedepth} and predict per-frame human instance segmentation masks $\{\Mv_{nt}\}$ with an image instance segmentation network, Mask2Former~\cite{cheng2022masked}. We adapt ZoeDepth for video-consistent depth estimation by choosing a per-video metric head from the majority vote of per-frame routers, for which we dub \zoedepthplus. While ZoeDepth claims to estimate metric depths, we observe a domain gap when inference on new datasets. Consequently, we only treat its output as up-to-affine depths that need to be further aligned with the metric scale. To aid our optimization with human awareness, we use camera-frame human meshes $\{\Vv_{nt}^\text{c}\}$ recovered by 4DHumans~\cite{goel2023humans} to introduce a metric prior.

\vspace{-0.5em}
\subsubsection{Calibrating Depth with Human Prior}
We calibrate the per-frame depths with human meshes in Human-aware Depth Calibration. This involves optimizing two parameters, a world scale $s$ and a world offset $o$, shared across all frames. During optimization, we linearly transform $\Dv_t$ to $\Dv'_t=s\Dv_t+o$ and unproject these depth maps to camera-frame point clouds $\{\Pv^\text{c}_t\}$ with $\Pv^\text{c}_t=\Pi^{-1}(\pv_t,\Dv'_t)$. Our intuition is to align the human point cloud $\Pv^\text{c}_{nt}=\Mv_{nt}\odot\Pv^\text{c}_{t}$ with the camera-frame human mesh vertices $\Vv_{nt}^\text{c}$ in terms of absolute depth and size. To achieve pixel-wise alignment, we use a depth term to pull points on the human point cloud toward their corresponding human mesh vertices along the z-axis
\begin{align}
    E_\text{depth} = \frac{\sum_{n,t} \norm{\Sv_{nt} \odot \left[z(\Vv_{nt}^\text{c})-z(\Pv^\text{c}_{nt})\right]}_2^2}{\sum_{n,t}\norm{\Sv_{nt}}_0},
\end{align}
where $\Sv_{nt}=\rho(\Vv_{nt}^\text{c})\cap\Mv_{nt}$ is the intersection of the rasterized human mesh mask $\rho(\Vv_{nt}^\text{c})$ and the instance segmentation mask $\Mv_{nt}$,  $z(\cdot)$ is the rasterized depth, and $\norm{\cdot}_0$ is the 0-norm indicating the number of non-zero pixels on a mask.

As the recovered human meshes can be noisy in depth but still have a stable body dimension, we also adopt a size term to leverage the relative position of mesh vertices
\begin{align}
    E_\text{dx} &= \frac{\sum_{n,t}\norm{\Delta_x(\Vv_{nt}^\text{c}, \Sv_{nt}) -\Delta_x(\Pv^\text{c}_{nt}, \Sv_{nt})}_2^2}{NT}.
\end{align}
We define $E_\text{dy}$ similarly as $E_\text{dx}$, where
\begin{align}
    \Delta_*(\Xv,\Yv) =& (\max_* - \min_*) \left[\Pi^{-1}(\Yv\odot\Pi(\Xv), z(\Xv))\right]
\end{align}
and $(\max_* - \min_*)$ denotes the difference between the maximum value and the minimum value on coordinate $*$.

Then we have the calibrated depths with optimization
\begin{align}
    (s^\text{m},o^\text{m}) &= \argmin_{s,o}{(E_\text{depth}+\lambda E_\text{size})}, \\
    \Dv_t^\text{m} &= s^\text{m}\Dv_t+o^\text{m},
\end{align}
where $E_\text{size} = E_\text{dx} + E_\text{dy}$, and $\lambda$ is a hyperparameter to balance two energy terms with a default value of 1.

\begin{figure}[t]
    \begin{center}
        \includegraphics[width=\linewidth]{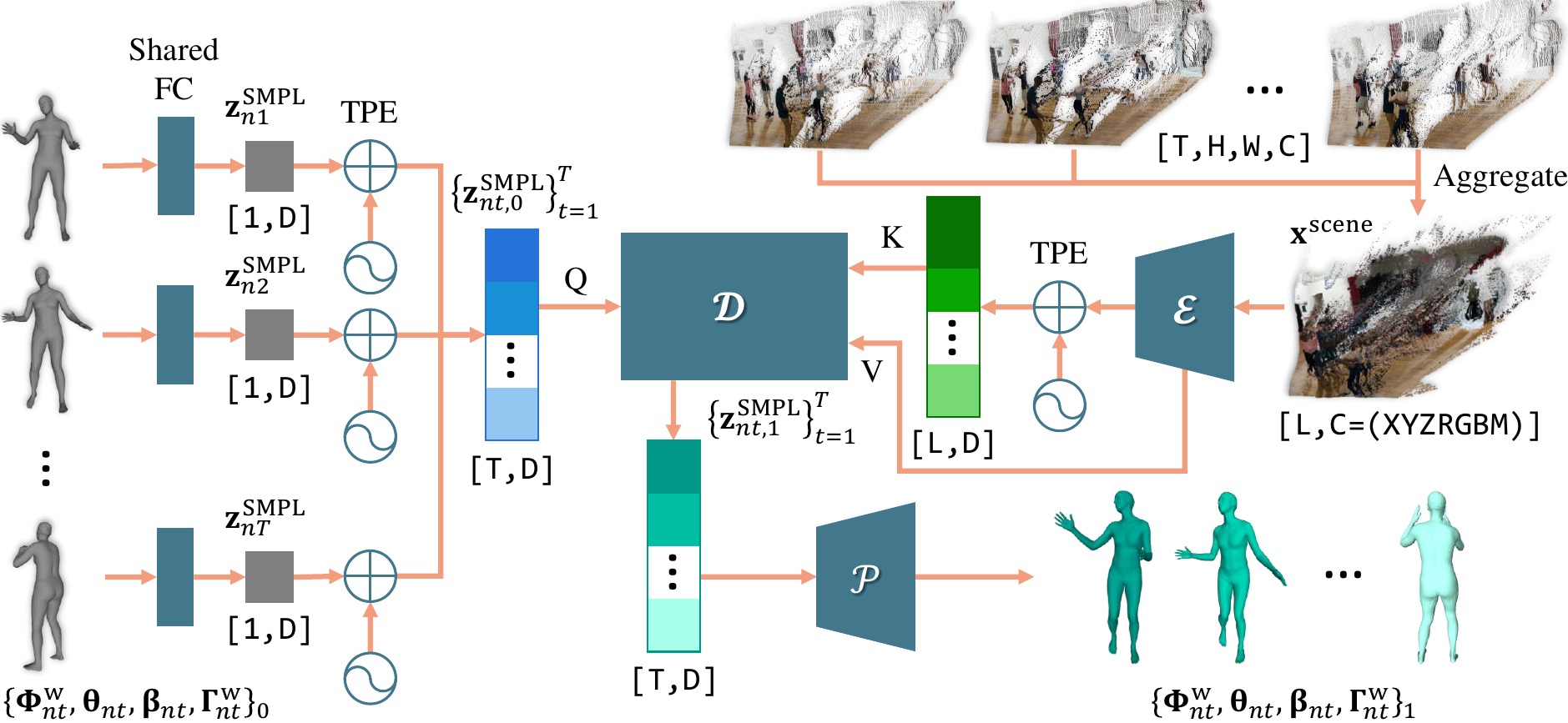}
        \caption{\textbf{The architecture of Scene-aware SMPL Denoiser.} World-frame noisy SMPL parameters $\{\Phiv_{nt}^\text{w}, \thetav_{nt}, \betav_{nt}, \Gammav_{nt}^\text{w}\}_0$ are first projected by a linear layer and summed with temporal positional embeddings (TPE) to get initial latent humans $\{\zv_{nt,0}^\text{SMPL}\}$. Per-frame point clouds are aggregated to $\xv_\text{scene}$ and encoded with the point encoder $\Ec$. Then we query the encoded scene $\Ec(\xv^\text{scene})$ with latent humans $\{\zv_{nt,0}^\text{SMPL}\}$ in the scene-conditioned denoiser $\Dc$ and feed the result $\{\zv_{nt,1}^\text{SMPL}\}$ to prediction heads $\{\Pc_{\Phiv},\Pc_{\thetav},\Pc_{\betav},\Pc_{\Gammav}\}$ to obtain denoised SMPL parameters $\{\Phiv_{nt}^\text{w}, \thetav_{nt}, \betav_{nt}, \Gammav_{nt}^\text{w}\}_1$.}
        \label{fig:denoiser}
    \end{center}
\vspace{-2.0em}
\end{figure}

\subsubsection{Disambiguating SLAM with Calibrated Depth}
While DROID-SLAM~\cite{teed2021droid} originally supports RGB-D input mode where the D channel stands for sensor depth, one cannot trivially access sensor depths from in-the-wild videos. Our insight is that an estimated absolute depth can be utilized as a depth prior, albeit noisy. So we combine the original RGB video and the calibrated depth as pseudo-RGB-D inputs $\{\Iv_t,\Dv_t^\text{m}\}$ to disambiguate depth and scale. Furthermore, we modify the cost function~\cref{eq:slam} to resolve the dynamic ambiguity by masking out dynamic foregrounds in confidence maps
\begin{align}
    \Sigma'_{ij} = \diag \wv'_{ij} = \diag \left((1-[\Mv_{i},\Mv_{j}])\odot \wv_{ij}\right),
\end{align}
where $\Mv_{i}=\bigcup_n{\Mv_{ni}}$ and $\Mv_{j}=\bigcup_n{\Mv_{nj}}$ are the union of all human instance masks on their corresponding frame, and $[\cdot,\cdot]$ is the concatenation operation. As a result, we obtain metric-scale camera poses $\{\Gv_t^\text{m}\}$ and metric-scale point clouds $\{\Pv_t^\text{wm}\}$ by disambiguating SLAM with calibrated metric depths
\begin{align}
    \{\Gv_t^\text{m},\dv_t^\text{m}\} &= \argmin_{\{\Gv_{ij}',\dv_{ij}'\}}E_{\Sigma'}, \\
    \Pv_t^\text{wm} &= \Gv_t^\text{m} \circ \Pi^{-1}(\pv_t, \frac{1}{\dv_t^\text{m}})).
\end{align}

\subsection{Scene-aware SMPL Denoising}
\label{subsec:smpl_denoising}
\subsubsection{Initializing Humans with Metric Cameras}
To put humans properly in the scene recovered by SLAM, we initialize them by transforming estimated camera-frame SMPL parameters $\{\Phiv_{nt}^\text{c}, \thetav_{nt}, \betav_{nt}, \Gammav_{nt}^\text{c}\}$ to the world frame with camera-to-world transforms $\{\Gv_t^\text{m}=[\Rv_t|\tv_t^\text{m}]\}$. 
Given the pelvis as the center of global orientation $\Phiv$, we have:
\begin{align}
    \Phiv_{nt}^\text{w}=\Rv_t\Phiv_{nt}^\text{c},\quad
    \Gammav_{nt}^\text{w}=\Rv_t(\Gammav_{nt}^\text{c}+\cv)+\tv_t^\text{m}-\cv,
\end{align}
where $\cv=\cv(\betav_{nt})$ is the pelvis location in the shape blend body mesh. Note that we do not need to introduce an extra camera scale as SLAHMR~\cite{ye2023slahmr} since the camera poses have already been in the metric scale. 
The root-relative poses $\thetav_{nt}$ and the shapes $\betav_{nt}$ stay unchanged as in the camera frame.
We denote the initialized and the denoised parameters with a suffix $0$ and $1$ respectively, \ie $\{\Phiv_{nt}^\text{w}, \thetav_{nt}, \betav_{nt}, \Gammav_{nt}^\text{w}\}_{0,1}$.

\subsubsection{Constraining Humans with Dynamic Scenes}
Different from existing works~\cite{kocabas2024pace,ye2023slahmr,yu2021movingcam,liu20214d} that incorporate energy terms in optimization to apply explicit scene constraints, we propose to learn implicit scene constraints with a Scene-aware SMPL denoiser shown in~\cref{fig:denoiser}. The noisy initial SMPL parameters $\{\Phiv_{nt}^\text{w}, \thetav_{nt}, \betav_{nt}, \Gammav_{nt}^\text{w}\}_0$ are first projected to a latent space, where it gets further updated by conditioning on implicit scene constraints
\begin{align}
    &\zv_{nt,0}^\text{SMPL} = \text{FC}\left([\Phiv_{nt,0}^\text{w}, \thetav_{nt,0}, \betav_{nt,0}, \Gammav_{nt,0}^\text{w}]\right)+\text{TPE}, \\
    \{&\zv_{nt,1}^\text{SMPL}\}_{t=1}^T = \Dc\left(\{\zv_{nt,0}^\text{SMPL}\}_{t=1}^T, \Ec(\xv^\text{scene})+\text{TPE}\right),
\end{align}
where FC is a shared linear layer, TPE is shared temporal positional embeddings, $\{\zv_{nt,*}^\text{SMPL}\}_{t=1}^T\in\Rb^{T \times D}$ is the $D$-dimensional latent for human $n$, and $\xv^\text{scene}\in\Rb^{L \times C}$ is the $C$-channel dynamic scene point clouds with a total number of points $L$. $\Ec$ and $\Dc$ refer to the scene encoder and the scene-conditioned denoiser, respectively. We set $C=7$ which is the concatenation of point coordinates $\{\Pv_t^\text{wm}\}$, colors $\{\Iv_t\}$, and estimated human semantic segmentation masks $\{\Mv_{t}=\bigcup_n{\Mv_{nt}}\}$. 
Following \cite{goel2023humans}, the updated latent are decoded with different prediction head $\Pc_{(\cdot)}$ to regress the residual for each SMPL parameter:
\begin{align}
    \Phiv_{nt,1}^\text{w} &= \Pc_{\Phiv}(\zv_{nt,1}^\text{SMPL}) \Phiv_{nt,0}^\text{w}, \\
    \thetav_{nt,1} &= \Pc_{\thetav}(\zv_{nt,1}^\text{SMPL}) \thetav_{nt,0}, \\
    \betav_{nt,1} &= \Pc_{\betav}(\zv_{nt,1}^\text{SMPL}) +  \betav_{nt,0}, \\
    \Gammav_{nt,1}^\text{w} &= \Pc_{\Gammav}(\zv_{nt,1}^\text{SMPL}) + \Gammav_{nt,0}^\text{w}.
\end{align}
We apply direct supervision on $\{\Phiv_{nt}^\text{w}, \thetav_{nt}, \betav_{nt}, \Gammav_{nt}^\text{w}\}_1$, which is common in the literature. Please see \supp for the details of the full training objectives.

%% file: sections/4_experiments.tex
\section{Experiments}
\label{sec:exp}
\subsection{Experimental Setting}
\minisection{Datasets.} We assess the performance of \nameMethod primarily for global human motion estimation but also report the accuracy of estimated camera trajectories.
Traditional video datasets in \hmr literature are typically captured by static cameras, \eg~\cite{ionescupapavaetal2014,Hassan2019prox,Mehta2018XNectRM,yoon2021humbi,joo2015panoptic}, hence not suitable for our purpose. 
Standard SLAM benchmarks such as \cite{sturm12iros,schops2019bad} do not meet our needs either as there is often no human moving in the scene.
We consider the following datasets. 

\noindent\textbf{\threedpw \cite{vonmarcard_eccv_2018_3dpw}} is an in-the-wild dataset captured with iPhones. The ground truth bodies are not in coherent world frames so we use it to supervise root relative poses and for evaluation.

\noindent\textbf{\egobody \cite{zhang2022egobody}} has ground-truth poses captured by multiple Kinects and egocentric-view sequences recorded by a head-mounted device, whose trajectories are further registered in the world space of Kinect array. 
We use it for training the SMPL denoiser in~\cref{subsec:smpl_denoising} and for evaluation (on both body and camera estimation).
For \hmr evaluation, unlike \cite{kocabas2024pace,ye2023slahmr} considering only the validation set, we additionally report results on its completely withheld test set.

\noindent\textbf{EMDB \cite{kaufmann2023emdb}} is a new dataset providing SMPL poses from IMU sensors and global camera trajectories.
We include it for training the SMPL denoiser to enrich the diversity and use the camera trajectories to evaluate the quality of \slam.

\minisection{Evaluation Metrics.}
For \hmr evaluation, we report common PA-MPJPE, which measures the quality of root-relative poses.
For datasets that have ground-truth poses in a world coordinate, we follow \cite{ye2023slahmr} and consider WA-MPJPE and FA-MPJPE. The former measures the error after aligning the entire trajectories of the prediction and ground truth with Procrustes Alignment \cite{gower1975generalized}, while the latter aligns only with the first frame. We also report acceleration errors.
For \slam, we consider absolute trajectory error (ATE) for camera trajectory evaluation as well as the threshold accuracy ($\delta_n$), the absolute relative error (REL), and the root mean squared error (RMSE) for scene depth evaluation~\cite{bhat2023zoedepth}.

\minisection{Implementation Details.} In Human-aware Depth Calibration, we use the L-BFGS algorithm with learning rate 1 to optimize for a maximum of 30 iterations. As for the Scene-aware SMPL Denoiser, we train it on the union of \threedpw-Train, EgoBody-Train, and EMDB for 100k steps with an AdamW optimizer, a batch size of 16, and a learning rate of 1e-5. For camera-frame SMPL ground truths like in \threedpw, we only incorporate body shapes $\betav$ and poses $\thetav$ in training. We train the denoising process by randomly sampling a temporal window size $T$ spanning 64 to 128 and inference with $T=100$. The scene-conditioned denoiser $\Dc$ is parameterized with a 6-layer Transformer Decoder. For the scene encoder $\Ec$, we consider ViT and SPVCNN in~\cref{tab:scene} and report results for SPVCNN in~\cref{tab:egobody,tab:3dpw}. Before inputting the world-frame noisy SMPL parameters to the denoiser, we first interpolate $\Phiv_{nt,0}^\text{w}$ and $\thetav_{nt,0}$ on $\text{SO}(3)$, $\betav_{nt,0}$ on $\Rb^{10}$, and $\Gammav_{nt,0}^\text{w}$ on $\Rb^{3}$ when there are missing observations.

\begin{table}[t]
\centering
\footnotesize
\setlength{\tabcolsep}{0.4em}
\adjustbox{width=\linewidth}
	{\begin{tabular}{ll|c}
        \toprule
            Camera Model & Human Model & PA $\downarrow$ \\
        \midrule
            DROID-SLAM~\cite{teed2021droid} & \slahmr~\cite{ye2023slahmr} w/ \phalpp & 55.9 \\
            DROID-SLAM~\cite{teed2021droid} & \slahmr~\cite{ye2023slahmr} w/ 4DHumans~\cite{goel2023humans} & 57.4 \\
            Human-aware Metric SLAM (ours) & 4DHumans~\cite{goel2023humans} & 52.9 \\
            \rowcolor{Gray} Human-aware Metric SLAM (ours) & Scene-aware SMPL Denoiser (ours) & \textbf{52.4} \\
        \bottomrule
	\end{tabular}}
\caption{\textbf{Comparison results on \threedpw-Test.} The row in gray is the full pipeline of \nameMethod. We abbreviate PA-MPJPE as PA, with the same below for FA-MPJPE (FA) and WA-MPJPE (WA).}
\label{tab:3dpw}
\vspace{-1.0em}
\end{table}

\begin{table*}[t]
\centering
\footnotesize
\setlength{\tabcolsep}{0.4em}
\adjustbox{width=\linewidth}
{
\begin{tabular}{c|ll|ccccc}
    \toprule
        Subset & Camera Model & Human Model & PA-MPJPE (mm) $\downarrow$ & FA-MPJPE (mm) $\downarrow$ & WA-MPJPE (mm) $\downarrow$ & Acc Error (mm/frame$^2$) $\downarrow$ & Runtime/100 imgs\\
    \midrule
        \multirow{6}{*}{Val} & - & GLAMR~\cite{yuan2021glamr} & 114.3 & 416.1 & 239.0 & 173.5 & 4 min \\
        & DROID-SLAM~\cite{teed2021droid} & PACE~\cite{kocabas2024pace} & 66.5 & 147.9 & 101.0 & \textbf{6.7} & 1 min \\
        & DROID-SLAM~\cite{teed2021droid} & \slahmr~\cite{ye2023slahmr} w/ \phalpp  & 79.1 & 141.1 & 101.2 & 25.8 & 40 min \\
        & DROID-SLAM~\cite{teed2021droid} & \slahmr~\cite{ye2023slahmr} w/ 4DHumans~\cite{goel2023humans} & 79.3 & 273.0 & 144.7 & 79.4 & 40 min \\
        & Human-aware Metric SLAM (ours) & 4DHumans~\cite{goel2023humans} & 73.0 & 164.4 & 106.7 & 127.0 & 5 min \\
        \rowcolor{Gray} & Human-aware Metric SLAM (ours) & Scene-aware SMPL Denoiser (ours) & \textbf{57.7} & \textbf{115.1} & \textbf{81.1} & 64.8 & 5 min \\
    \midrule
        \multirow{5}{*}{Test} & - & GLAMR \cite{yuan2021glamr} & 112.8 & 351.4 & 216.3 & 105.9 & 4 min \\
        & DROID-SLAM~\cite{teed2021droid} & \slahmr~\cite{ye2023slahmr} w/ \phalpp & 63.1 & 163.9 & 99.4 & \textbf{31.7} & 40 min \\
        & DROID-SLAM~\cite{teed2021droid} & \slahmr~\cite{ye2023slahmr} w/ 4DHumans~\cite{goel2023humans} & 69.3 & 185.8 & 113.0 & 45.7 & 40 min \\
        & Human-aware Metric SLAM (ours) & 4DHumans~\cite{goel2023humans} & 75.4 & 160.0 & 108.1 & 138.8 & 5 min \\
        \rowcolor{Gray} & Human-aware Metric SLAM (ours) & Scene-aware SMPL Denoiser (ours) & \textbf{61.3} & \textbf{122.1} & \textbf{84.6} & 69.4 & 5 min \\
    \bottomrule
\end{tabular}
}
\vspace{-0.5em}
\caption{\textbf{Comparison results with state-of-the-art approaches on \egobody.} The row in gray is the full pipeline of \nameMethod.}
\label{tab:egobody}
\vspace{-1.3em}
\end{table*}

\begin{table}[ht]
\centering
\footnotesize
\setlength{\tabcolsep}{0.4em}
\adjustbox{width=\linewidth}{
\begin{tabular}{ccc|cccc|c}
    \toprule
        \multirow{2}{*}{RGB} & \multirow{2}{*}{Depth} & \multirow{2}{*}{Mask} & \multicolumn{4}{c|}{EgoBody} & EMDB \\
        &  &  & ATE $\downarrow$ & $\delta_1 \uparrow$ & REL $\downarrow$ & RMSE $\downarrow$ & ATE $\downarrow$ \\
    \midrule
        \cmark & \xmark & \xmark & 80.9 & 0.085 & 14.590 & 1617.361 & 400.3 \\
        \cmark & \xmark & Mask2Former~\cite{cheng2022masked} & 81.6 & 0.063 & 8.530 & 1009.127 & 385.8 \\
        \cmark & \zoedepthplus & \xmark & 35.0 & 0.562 & 0.308 & 15.360 & 456.8 \\
        \cmark & \zoedepthplus & Mask2Former~\cite{cheng2022masked} & 28.6 & 0.564 & 0.307 & 10.852 & 389.6 \\
        \rowcolor{Gray} \cmark & \zoedepthplus + Cal. & Mask2Former~\cite{cheng2022masked} & \textbf{26.4} & \textbf{0.797} & \textbf{0.274} & \textbf{10.452} & \textbf{107.0} \\
    \bottomrule
\end{tabular}
}
\vspace{-0.5em}
\caption{\textbf{Ablation study for SLAM configurations in terms of optimized camera trajectories and scene depths.}
\zoedepthplus denotes our video-adapted ZoeDepth~\cite{bhat2023zoedepth}.
}
\label{tab:slam}
\vspace{-0.5em}
\end{table}

\begin{table}[ht]
\centering
\footnotesize
\setlength{\tabcolsep}{0.4em}
\adjustbox{width=\linewidth}{
\begin{tabular}{llccc|cccc}
    \toprule
        Stage & Backbone & RGB & XYZ & Mask & PA $\downarrow$ & FA $\downarrow$ & WA $\downarrow$ & Acc Error $\downarrow$ \\
    \midrule
        Init. & - & \xmark & \xmark & \xmark & 73.7 & 120.8 & 93.1 & 127.1 \\
        Pred. & - & \xmark & \xmark & \xmark & 63.3 & 98.8 & 77.2 & 75.2 \\
        Pred. & ViT~\cite{dosovitskiy2020image} & \cmark & \xmark & \xmark & 63.9 & 94.9 & 76.7 & \textbf{43.3}  \\
        Pred. & ViT~\cite{dosovitskiy2020image} & \cmark & \cmark & \xmark & 64.5 & 97.3 & 77.7 & 45.6 \\
        Pred. & ViT~\cite{dosovitskiy2020image} & \cmark & \xmark & \cmark & 66.8 & 96.5 & 78.6 & 44.7  \\
        Pred. & ViT~\cite{dosovitskiy2020image} & \cmark & \cmark & \cmark & 69.3 & 100.9 & 82.0 & 46.4 \\
        Pred. & SPVCNN~\cite{tang2020searching} & \xmark & \cmark & \xmark & 62.9 & 95.1 & 76.0 & 72.6 \\
        Pred. & SPVCNN~\cite{tang2020searching} & \cmark & \cmark & \xmark & \textbf{61.0} & 93.4 & 74.3 & 67.7 \\
        Pred. & SPVCNN~\cite{tang2020searching} & \xmark & \cmark & \cmark & 62.0 & 93.9 & 75.3 & 69.9 \\
        \rowcolor{Gray} Pred. & SPVCNN~\cite{tang2020searching} & \cmark & \cmark & \cmark & 61.3 & \textbf{91.9} & \textbf{73.6} & 64.8 \\
    \bottomrule
\end{tabular}
}
\vspace{-0.5em}
\caption{\textbf{Ablation study for different scene encoders and features regarding world-frame HMR.} Init.~and Pred.~refer to before and after SMPL denoising, respectively.}
\label{tab:scene}
\vspace{-1.3em}
\end{table}

\subsection{Comparison Results} \label{subsec:quantitative}
We first evaluate the estimated local poses with PA-MPJPE on \threedpw, which is common in the literature.
In~\cref{tab:3dpw}, we show that placing the bodies from \fourdhumans already leads to lower error than \slahmr. Passing them through the denoiser further reduces the error.
We note that PA-MPJPE only measures local pose accuracy not the quality of global trajectories.
Since \threedpw does not support any world metrics, 
\cref{tab:3dpw} only aims to show that \nameMethod produces reasonable local poses on a common dataset. 

Next, we assess the quality of global motion estimation, which is essentially a more challenging task.
\cref{tab:egobody} shows the results on \egobody.
Note that current optimization-based methods \cite{ye2023slahmr,kocabas2024pace} report the error of the validation set. 
For fairness and completeness, we report results on both validation and test sets and run state-of-the-art methods on the test set when the code is available.
In~\cref{tab:egobody}, we see that the proposed \nameMethod has the overall lowest PA-MPJPE, FA-MPJPE, and WA-MPJPE (gray rows).
Comparing it with the row above (\fourdhumans) confirms the benefit of our scene-conditioned denoiser. 
For a fair comparison, we also initialize the global optimization of \slahmr with \fourdhumans, which is more accurate than \phalpp in \slahmr, but we do not observe improvement.
Notably, despite the concurrent work PACE \cite{kocabas2024pace} has a tightly integrated SLAM and body fitting objective, it still uses native DROID-SLAM to initialize the camera parameters like \slahmr does. 
This is arguably sub-optimal as the initialization is not aware of body information, which can lead to errors that cannot be corrected in the global optimization stage. 
Consequently, it also has higher world-space errors.
Optimization methods often employ a zero velocity term to smooth out human motion, which explains the lower acceleration error.
However, we do not observe a big difference in jittery between our results. Please refer to \supp for more details.
\subsection{Ablation Study}
\label{subsec:ablation}
We ablate the design choices in \nameMethod. 
In~\cref{tab:slam}, we evaluate SLAM-optimized camera trajectories and scene depths with \egobody and EMDB. 
We see that directly including un-calibrated monocular depths does not guarantee more accurate estimations (3\textsuperscript{rd} vs.~1\textsuperscript{st} and 4\textsuperscript{th} vs.~2\textsuperscript{nd} row).
Precluding the dynamic foreground pixels with Mask2Former~\cite{cheng2022masked} generally improves performance. 
We empirically find that our depth calibration with human prior works the best when using it with foreground masking, which has the lowest error in both datasets. More SLAM evaluation and discussion can be found in \supp

In~\cref{tab:scene}, we verify the benefit of scene conditioning for the SMPL denoiser. 
We train it with EgoBody-train in different conditioning schemes and report the $T=32$ results on EgoBody-val.
First, placing the predicted bodies from 4DHuman in the global space directly with estimated camera extrinsics has the highest error (1\textsuperscript{st} row). 
When conditioning on a constant zero tensor, the denoiser behaves like a motion prior and reduces the error (2\textsuperscript{nd} row). 
To encode the appearance and geometry information of the scene, we consider ViT \cite{dosovitskiy2020image} or SPVCNN \cite{tang2020searching} as the encoder $\Ec$ and try varied combinations of appearance features (RGB), geometry features (XYZ) and aggregated subject masks (Mask).
When using ViT to encode the scene, adding XYZ features or masks does not reduce the error. 
In contrast, when using SPVCNN, adding RGB information or conditioning on masks does improve performance. 
Overall, SPVCNN yields lower errors than ViT and enabling all conditioning leads to the lowest world-space error measure. 

\begin{figure*}[ht]
    \begin{center}
        \includegraphics[width=\linewidth]{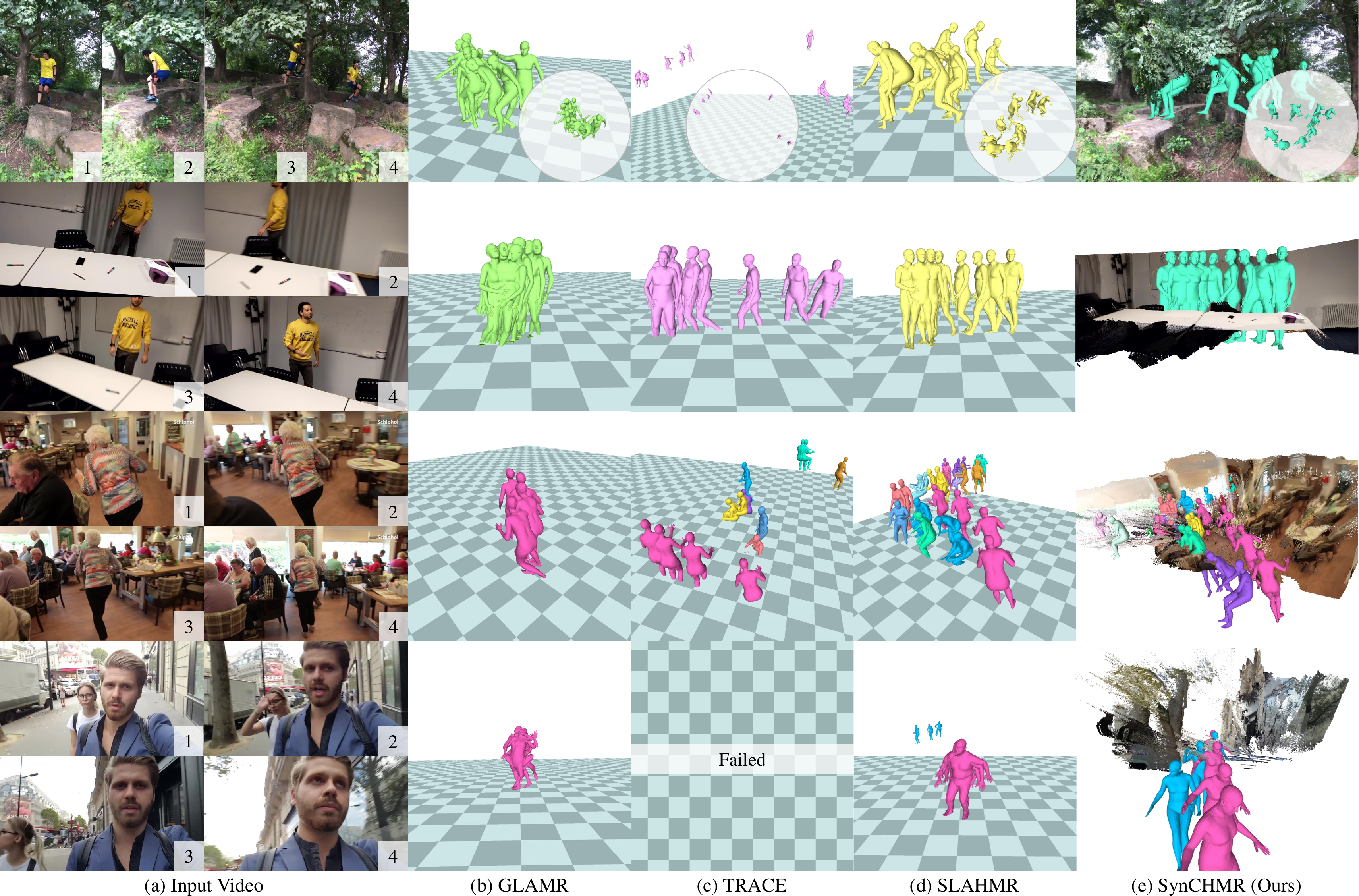}
        \caption{\textbf{Qualitative comparison among \worldspacehmr approaches.} We show (b) GLAMR~\cite{yuan2021glamr} and (c) TRACE~\cite{sun2023trace} results with their pre-defined ground planes, (d) SLAHMR~\cite{ye2023slahmr} outputs with its estimated ground plane, and (e) our \synchmr outputs with dense scenes. In the first row, we also demonstrate top-view human trajectories within circles. See supplementary for video results.}
        \label{fig:qual}
        \vspace{-2.0em}
    \end{center}
\end{figure*}

\subsection{Qualitative Analysis and Discussion}
In the first two rows of~\cref{fig:qual}, we visualize the results of \threedpw and \egobody in a global space.
Despite occlusions, our \nameMethod estimates human meshes reliably and places them in a dense scene point cloud, whereas the scenes in GLAMR~\cite{yuan2021glamr} and \slahmr~\cite{ye2023slahmr} consist of only a simple ground plane.
Applying scene constraints with such an overly simplified scene can result in erroneous estimation, \eg, incorrect human trajectories as shown in the top view of the 1\textsuperscript{st} row, and the vertically shortened human bodies in the 4\textsuperscript{th} row of (d). Note that since TRACE~\cite{sun2023trace} is scene agnostic, the ground plane in (c) is only for visualization, not necessarily indicating scene penetration.

We also test on more in-the-wild DAVIS~\cite{Perazzi2016} videos containing human subjects.
Since DAVIS provides no ground-truth human meshes nor camera trajectories, we show only the visual comparison. 
The 3\textsuperscript{rd} row shows that we can handle multi-person cases as well as \slahmr, while GLAMR often fails when multiple humans and dynamic cameras both occur.
In a challenging scenario where the subject is taking selfies (the 4\textsuperscript{th} row), both GLAMR and \slahmr are confused by the foreground human dominating the frames and reconstruct an almost static global trajectory, failing to disentangle the camera and the human motions due to the dynamic ambiguity. 
TRACE fails to produce results due to severe frame truncation.
In contrast, \nameMethod still successfully provides reasonable trajectories.

%% file: sections/5_limitation.tex
\section{Limitation Discussion}
As \nameMethod focuses on disentangling camera and human movements, we follow \slahmr to approximate the focal length as $\frac{W+H}{2}$. 
When the subject has a shape that the body model cannot explain well, \eg, children or obese people, 
calibrating depth with the estimated bodies is less ideal.
As we develop and validate \nameMethod on real videos, its accuracy on composed or generated videos remains an open question.
Finally, since \nameMethod handles dynamic scenes with moving subjects, it does not require an \textit{a priori} scanned static scene. This opens up new challenges, such as incorporating dynamic point clouds as scene constraints.



%% file: sections/6_conclusion.tex
\section{Conclusion}
We present \nameMethod, a method that reconstructs camera trajectories, human bodies, and dense scenes from in-the-wild videos all in one global coordinate. \nameMethod has two core innovations.
First, it leverages monocular depth estimation and uses the dimension and location of human meshes to calibrate the range of depth. 
This allows \slam to better resolve the inherent scale ambiguity problem as shown in the experiment. 
Second, we train a data-driven motion denoiser and condition it with the scene in the same global coordinate, which is the first such scene-conditioned motion prior. 
Combining the two, the full \nameMethod pipeline uses human bodies to improve \slam, and the better estimated scene and camera trajectory, in turn, provide better constraints for feed-forward human motion denoising.
It achieves SOTA results on common benchmarks compared with existing optimization-based approaches.


%% file: sections/7_acknowledgment.tex
\section*{Acknowledgment}
We appreciate constructive comments from Duygu Ceylan. This project was partially supported by the NIH under contracts R01GM134020 and P41GM103712, and by the NSF under contracts DBI-1949629, DBI-2238093, IIS-2007595, IIS-2211597, and MCB-2205148.

%% file: sections/X_suppl.tex
\clearpage
\setcounter{page}{1}
\maketitlesupplementary

In this document, we provide additional technical details, more ablation studies, and more discussions. We refer the readers to the accompanying webpage for video results.
\section{\nameMethod Setting vs.~Prior Work}
\label{supsec:related}
We compare the setup of recent \worldspacehmr methods that handle dynamic cameras in~\cref{tab:literature}. 
Methods that estimate world-frame body parameters through learning-based approaches often ignore the camera at test time~\cite{sun2023trace,yu2021movingcam,li2022dnd}. 
On the other hand, optimization approaches need to estimate the camera at test time to fit to the detected 2D joint key points~\cite{yuan2021glamr,liu20214d,henning2022bodyslam,saini2023smartmocap,kocabas2024pace,ye2023slahmr}, and
we have discussed the downsides of their camera estimation approaches in Sec.~\ref{sec:related} of the main paper.
It is still worth noting that none of these methods reconstruct dense scene point clouds, except Liu \etal \cite{liu20214d}, who adopt COLMAP \cite{schoenberger2016sfm} for this purpose.
However, since COLMAP is not robust enough for in-the-wild videos, they demonstrate results only on sequences acquired in a controlled capture settings.
In stark contrast, \nameMethod is designed to work on casual videos.
It does not assume the scene is a ground plane as in \cite{ye2023slahmr,kocabas2024pace} or is scanned \textit{a priori} as in \cite{huang2022rich,Hassan2019prox}.
It has a light-weight setup but it reconstructs the most information -- human meshes, camera trajectory, and dense scene, all in one coherent global space.

\section{Training Objectives for SMPL Denoiser}
\label{supsec:detail}
We consider a simple linear layer for each prediction head and parameterize $\Phiv$ and $\thetav$ predictions as quaternions. 
Specifically, $\Pc_{\Phiv}:\Rb^D\rightarrow\Rb^{4}$, $\Pc_{\thetav}:\Rb^D\rightarrow\Rb^{J \times 4}$, $\Pc_{\betav}:\Rb^D\rightarrow\Rb^{10}$, and $\Pc_{\Gammav}:\Rb^D\rightarrow\Rb^{3}$, where $J$ denotes the number of joints. Then we apply direct supervision of SMPL parameters to the predictions
\begin{align*}
    \Lc_{\Phiv} &= 1 - \lVert \text{q}(\Phiv){\text{q}(\Phiv^*)^\top} \rVert_1, \\
    \Lc_{\thetav} &= 1 - \lVert \text{q}(\thetav){\text{q}(\thetav^*)^\top} \rVert_1, \\
    \Lc_{\betav} &= \lVert \betav - {\betav}^* \rVert_1, \\
    \Lc_{\Gammav} &= \lVert \Gammav - {\Gammav}^* \rVert_1,
\end{align*}
where $\text{q}(\cdot)$ stands for the quaternion representations and superscript $^*$ denotes the ground truth. Following~\cite{goel2023humans}, we also introduce a discriminator $\Cc$ to ensure the per-frame predictions are valid
\begin{align*}
    \Lc_{\Cc} &= \lVert \boldsymbol{1} - \Cc(\thetav,\betav) \rVert_2^2.
\end{align*}
The parameters are first factorized into (i) body pose parameters, (ii) shape parameters, and (iii) per-part relative rotations and classified by the discriminator to be fake (0) or real (1). To account for human motion, we further supervise the velocities and accelerations of human joints
\begin{align*}
    \Lc_{\dot\Jv} &= \left\lVert {\lVert\dot\Jv\rVert_2 - \lVert\dot\Jv^*\rVert_2} \right\rVert_1, \\
    \Lc_{\ddot\Jv} &= \left\lVert {\lVert\ddot\Jv\rVert_2 - \lVert\ddot\Jv^*\rVert_2} \right\rVert_1,
\end{align*}
where $\Jv$ are SMPL regressed joint locations.

\section{\slam Evaluation}
\begin{figure*}[ht]
    \begin{center}
        \includegraphics[width=0.95\linewidth]{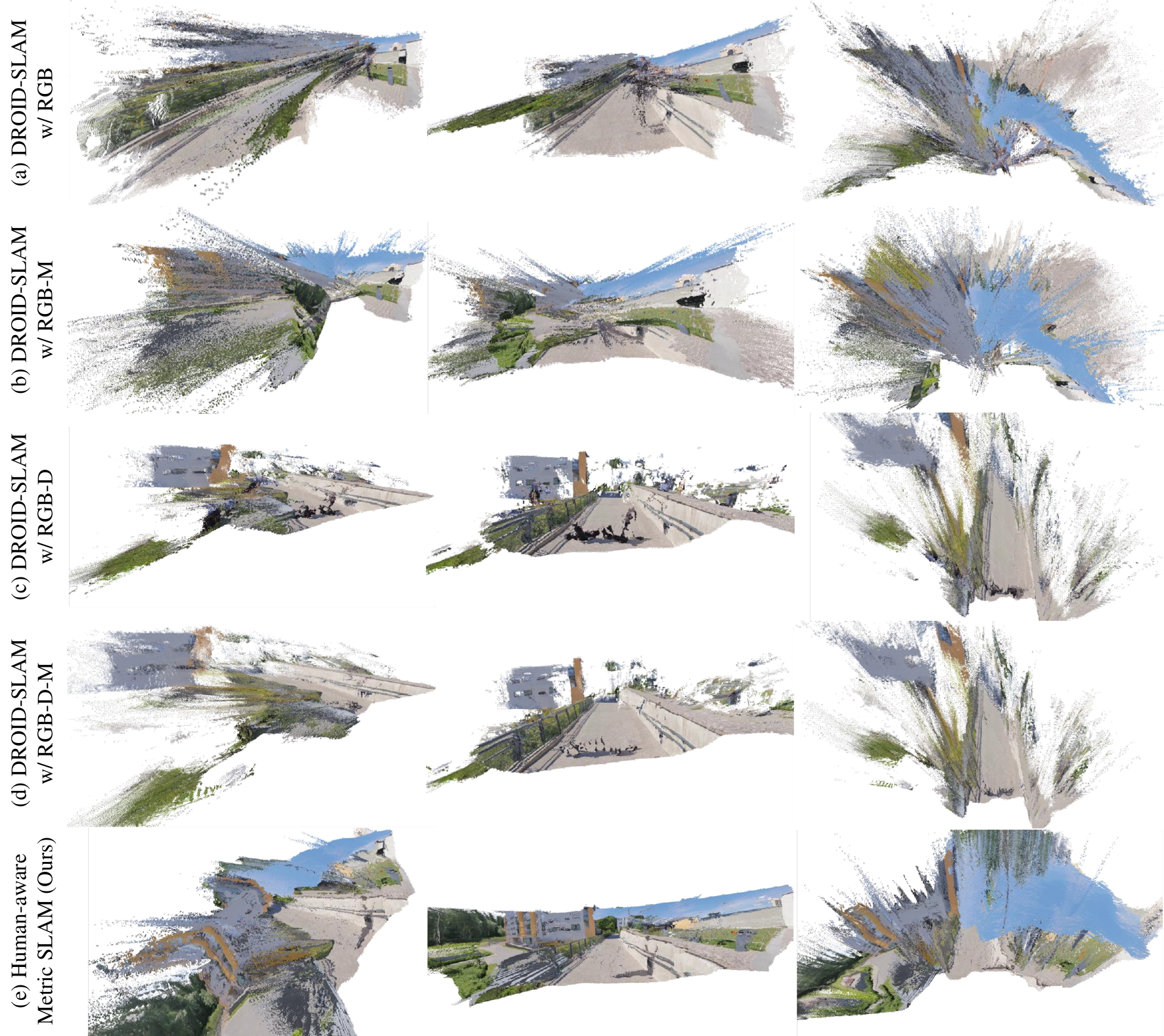}
    \end{center}
    \vspace{-1.0em}
    \caption{\textbf{Qualitative comparisons of the parkour sequence from DAVIS~\cite{Perazzi2016}.} (a) naive DROID-SLAM~\cite{teed2021droid} reconstructed point cloud with RGB input; (b) DROID-SLAM reconstructed point cloud with RGB input, where the foreground humans are masked out by an instance segmentation method Mask2Former~\cite{cheng2022masked}; (c) DROID-SLAM reconstructed point cloud with RGB-D input, where the depth channel is from ZoeDepth~\cite{bhat2023zoedepth} estimations, the same below; (d) DROID-SLAM reconstructed point cloud with RGB-D and instance segmentation mask inputs (e) our proposed Human-aware Metric SLAM reconstructed point cloud. Please see the webpage for video results.}
    \label{fig:parkour}
\vspace{-2.0em}
\end{figure*}

\minisection{Qualitative ablation study.}
\label{supsec:ablation}
In~\cref{tab:slam} of the main paper we quantitatively analyze the contribution of each design choice in our human-aware \slam; here we provide visual examples.
In~\cref{fig:parkour}, we show the results where we gradually add each design choice as stronger priors to the native visual SLAM.
Merely using RGB inputs in~\cref{fig:parkour}(a), naive DROID-SLAM~\cite{teed2021droid} fails in capturing the geometry structure of the scene. This results in a back-folded corridor, which is far from reasonable. The dynamic human also confuses the SLAM model, leading to a messy human point cloud in the center and everything else surrounding it in a circular shape. Masking out the human in~\cref{fig:parkour}(b) only removes the messy human point cloud but still produces a broken geometry since the depth ambiguity remains. An extra estimated depth channel in~\cref{fig:parkour}(c)(d) helps to resolve the depth ambiguity and correct the scene geometry. However, as we filter out points with epipolar inconsistency, the resulting point cloud is rather sparse. This indicates depth estimation with ZoeDepth~\cite{bhat2023zoedepth} does not guarantee each point has a consistent location across different frames, and SLAM fails to correct this error. Finally, our Human-aware Metric SLAM in~\cref{fig:parkour}(e) is able to output a dense point cloud. This reflects the success in finding more points with consistent 3D locations. As the scene reconstruction depends on camera pose estimation in SLAM, our pipeline potentially produces more accurate camera poses.

\minisection{Results on TUM-RGBD dataset.}
\cref{tab:slam} of the main paper considers \hmr datasets that provide ground truth camera trajectories.
Here, we report the results on a common \slam benchmark TUM-RGBD~\cite{sturm12iros}.
Since it does not contain humans in the scene, we can only apply our adapted video-consistent ZoeDepth~\cite{bhat2023zoedepth}, namely ZoeDepth$^+$, without calibrating the scales. 
In~\cref{tab:tum}, we see that this depth-augmented version yields an average lower error than the original DROID-SLAM. 
This suggests that despite the unknown scale, estimated monocular depth still provides prior information to better reason about camera trajectories.
One can see this as a byproduct of \nameMethod.

\begin{table*}[ht]
    \centering
    \footnotesize
    \begin{tabular}{r|c|c|c}
    \toprule
    Methods & Test-time Camera Estimation & Test-time Scene Representation & World-frame SMPL Params.~Estimation \\ 
    \midrule
    Yu \etal~\cite{yu2021movingcam}     & no camera estimation  & manually created shape primitives & RL-based \\ \hline
    TRACE \cite{sun2023trace}     & no camera estimation & no scene &feed-forward \\ \hline
    D\&D \cite{li2022dnd}     & estimated acceleration and angular velocity & ground plane  &feed-forward \\ \hline
    Liu \etal~\cite{liu20214d}     & COLMAP \cite{schoenberger2016sfm}& dense point cloud & optimization \\ \hline
    
    \multirow{2}{*}{GLAMR \cite{yuan2021glamr}}     & difference between the root transformations & \multirow{2}{*}{ground plane} & \multirow{2}{*}{optimization} \\ 
    & in the camera space and world space& & \\ \hline
    SmartMocap \cite{saini2023smartmocap}  & jointly solved with body params.; target: \textit{only} body kpts. & no scene &optimization\\ \hline 
    BodySLAM \cite{henning2022bodyslam} & jointly solved with body params.  & no scene &optimization\\ \cline{1-1} \cline{3-4}
    PACE \cite{kocabas2024pace}      & target: scene kpts and body kpts & ground plane & optimization \\ \hline
    \slahmr \cite{ye2023slahmr}  & DROID-SLAM \cite{teed2021droid} where humans are \textit{not} excluded  & ground plane &optimization \\ \hline
    \rowcolor{Gray}\nameMethod (ours) & human-aware metric SLAM (\cref{subsec:human-aware-slam}) & dense point cloud & scene-aware SMPL denoising (\cref{subsec:smpl_denoising}) \\
    \bottomrule
    \end{tabular}
    \caption{\textbf{Comparison of methods that reconstruct humans in a global space from a video filmed by a dynamic camera.}}
    \label{tab:literature}
\end{table*}

\begin{table*}[t]
    \centering
    \setlength{\tabcolsep}{0.4em}
    \adjustbox{width=0.75\linewidth}{
    \begin{tabular}{ccc|ccccccccc|c}
        \toprule
        RGB & Depth & Mask & 360 &     desk &    desk2 &    floor &     plant &     room &      rpy &     teddy &      xyz &       avg \\
        \midrule
        \cmark & \xmark & \xmark & 162.3   & 75.1  & 682.8  & 54.2   & 257.7   & 930.5  & 40.4  & 480.0   & 16.4  & 340.2   \\
        \cmark & ZoeDepth$^+$ & \xmark & 101.3   & 153.9  &  75.6 & 817.4  & 219.4    & 96.3  & 32.6 & 201.2  & 21.8  & 223.4 \\
        \bottomrule
    \end{tabular}
    }
    \caption{\textbf{Comparison between native DROID-SLAM (top) and our depth-augmented version (bottom) on TUM-RGBD \cite{sturm12iros}.}}
    \label{tab:tum}
\end{table*}

\begin{figure*}[t]
    \begin{center}
        \includegraphics[width=\linewidth]{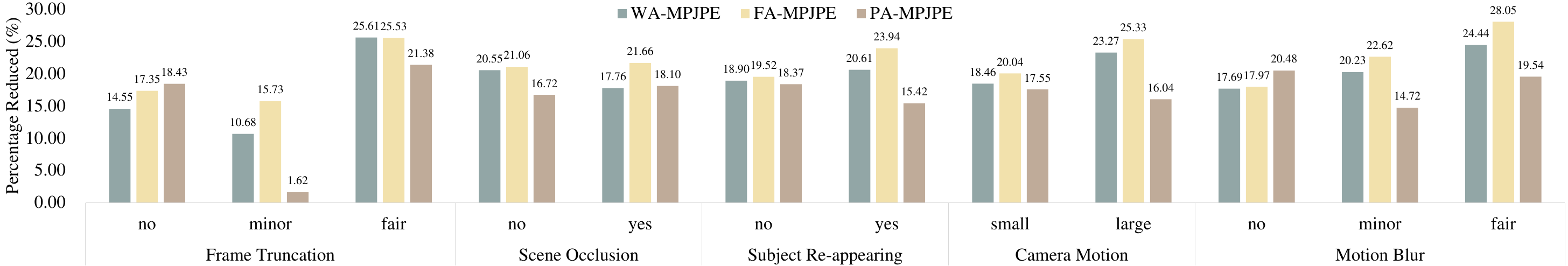}
    \end{center}
    \caption{\textbf{Percentage of MPJPEs reduced by Scene-aware SMPL Denoiser in videos with varied attributes.} The larger the better ($\uparrow$).}
    \label{fig:denoiser_analysis}
    \begin{center}
        \includegraphics[width=\linewidth]{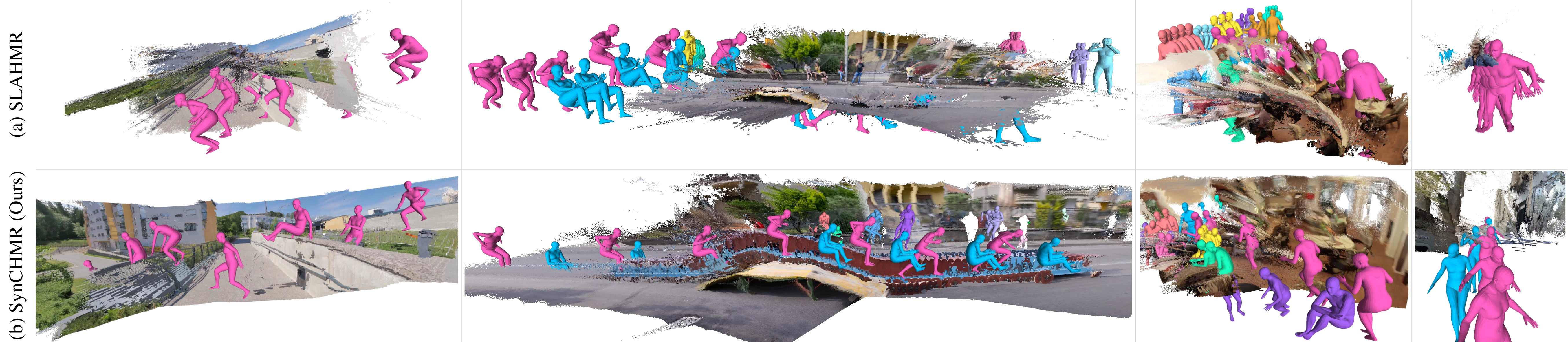}
    \end{center}
    \caption{\textbf{SLAHMR and SynCHMR human meshes integrated with static scenes.} Video visualizations are included in the webpage.}
    \label{fig:qualitative_comparison}
\end{figure*}

\section{\hmr Evaluation}
\minisection{Qualitative comparison.}
In~\cref{fig:qualitative_comparison}, we compare the estimated human body meshes and scene point clouds of (a) \slahmr~\cite{ye2023slahmr} and (b) our \synchmr. We observe incompatible scales and structures in \slahmr visualizations.
This can be the reason why \slahmr uses a ground plane instead of point clouds in the global refinement stage. 

\minisection{SMPL denoiser analysis.}
To better understand the impact of our scene-aware SMPL denoiser, we annotate the test set of EgoBody~\cite{zhang2022egobody} with 5 attributes: frame truncation, scene occlusion, subject reappearing, camera motion, and motion blur. 
In~\cref{fig:denoiser_analysis}, we plot the amount of error \textit{reduced} by SMPL denoiser in these attributes.
First, it confirms that the denoiser always brings improvement as there are no negative numbers. 
Second, we identify truncation, large camera motion, and motion blur as three primary scenarios where the denoiser helps greatly, as we see noticeable upward trends for them.
The underlying mechanism might be our SMPL denoiser captures more comprehensive scene information with dynamic scene modeling, which is beneficial in these situations where single-frame observations are bad and one needs to rely on cross-frame clues.

\minisection{Runtime analysis.}
We report the runtime of our \synchmr along with state-of-the-art models in~\cref{tab:egobody}. Note that the runtime for PACE~\cite{kocabas2024pace} does not include camera-frame initialization with HybrIK~\cite{li2021hybrik}.
To integrate per-frame human bodies into a smooth motion, \slahmr~\cite{ye2023slahmr} employs a HuMoR-like motion prior, which is slow due to its auto-regressive nature. 
PACE~\cite{kocabas2024pace} improves this by proposing a parallel motion prior.
Similarly, while adding in scene awareness, our feed-forward SMPL Denoiser also benefits from the parallel inference of the Transformer architecture.